\documentclass[10pt,twocolumn,letterpaper]{article}

\usepackage{cvpr}              % To produce the CAMERA-READY version
\usepackage[dvipsnames]{xcolor}

\definecolor{cvprblue}{rgb}{0.21,0.49,0.74}
\usepackage[pagebackref,breaklinks,colorlinks,citecolor=cvprblue]{hyperref}
\usepackage[capitalize]{cleveref}
\crefname{section}{Sec.}{Secs.}
\Crefname{section}{Section}{Sections}
\Crefname{table}{Table}{Tables}
\crefname{table}{Tab.}{Tabs.}

\usepackage{enumitem} % reduce space in enum
\usepackage{algorithm}
\usepackage{algorithmic}
\usepackage{booktabs}
\usepackage{multirow}
\usepackage{rotating}
\usepackage{adjustbox}
\def\eg{\emph{e.g}\onedot} 

\def\ie{\emph{i.e}\onedot} 

\usepackage{mathtools}
\usepackage{textcomp}
\usepackage{gensymb}
\usepackage[dvipsnames]{xcolor}

\newcommand\ours{MCLoc}
\newcommand{\gtr}[1]{}
\newcommand{\TSr}[1]{}

\newcommand{\myparagraph}[1]{\vspace{4pt}\noindent\textbf{#1.}}

\def\paperID{11925} % *** Enter the Paper ID here
\def\confName{CVPR}
\def\confYear{2024}

\title{The Unreasonable Effectiveness of Pre-Trained Features for Camera Pose Refinement}

\author{Gabriele Trivigno$^{1}$
\quad
Carlo Masone$^{1}$
\quad
Barbara Caputo$^{1}$
\quad
Torsten Sattler$^{2}$\\
$^{1}$ Politecnico di Torino \\
$^{2}$ Czech Institute of Informatics, Robotics and Cybernetics, Czech Technical University in Prague \\
{\tt\small \{gabriele.trivigno,carlo.masone,barbara.caputo\}@polito.it \, torsten.sattler@cvut.cz}
}

\begin{document}
\maketitle
\begin{abstract}
Pose refinement is an interesting and practically relevant research direction. Pose refinement can be used to (1) obtain a more accurate pose estimate from an initial prior (e.g., from retrieval), (2) as pre-processing, i.e., to provide a better starting point to a more expensive pose estimator, (3) as post-processing of a more accurate localizer. Existing approaches focus on learning features / scene representations for the pose refinement task. This involves training an implicit scene representation or learning features while optimizing a camera pose-based loss. A natural question is whether training specific features / representations is truly necessary or whether similar results can be already achieved with more generic features. In this work, we present a simple approach that combines pre-trained features with a particle filter and a renderable representation of the scene. Despite its simplicity, it achieves state-of-the-art results, demonstrating that one can easily build a pose refiner without the need for specific training. The code is  at {\small{\url{https://github.com/ga1i13o/mcloc_poseref}}}
\end{abstract}

\section{Introduction}
\label{sec:intro}

Visual localization estimates the position and the rotation of a camera in a given scene. It is essential in a wide range of applications such a Simultaneous Localization and Mapping (SLAM)~\cite{durrant2006simultaneous,bailey2006simultaneous2}, Structure-from-Motion (SfM)~\cite{schoenberger2016sfm,schoenberger2016mvs}, autonomous navigation~\cite{naseer2018robust,cummins2010fab}, robotics~\cite{fox1999monte,fox1999markov}, and Augmented-Virtual Reality (AR/VR)~\cite{pons2023interaction,guzov2021human}.\\
State-of-the-art methods follow a structure-based approach~\cite{sarlin2019coarse,sarlin2018leveraging} where a 3D map of the scene is available and a query image is localized against it by deriving 2D-3D matches.
Such 2D-3D correspondences are obtained by matching local features~\cite{revaud2019r2d2,dusmanu2019d2,superpoint} between the query image and the 3D points in the map, typically an SfM~\cite{schoenberger2016mvs,schoenberger2016sfm} sparse point cloud. These matches are used to estimate the camera pose with minimal solvers~\cite{hartley2003multiple,persson2018lambda} integrated with robust optimization~\cite{fischler1981random,chum2003locally}.
Generating the point cloud via SfM involves local feature detection and description on a set of reference images, feature matching, and triangulation of image points that are co-visible in several images~\cite{schoenberger2016sfm,hartley2003multiple}. The resulting 3D points are then associated with visual descriptors from the reference images. \\
While SfM-based point clouds enable robust and accurate localization~\cite{sarlin2019coarse,sattler2018benchmarking,sarlin2020superglue,sun2021loftr}, they remain \textit{unflexible} as they are tied to the specific features used for the reconstruction and their use is limited to the localization task~\cite{newcombe2011denslam,bloesh2019meshslam,Panek2022ECCV}. 
\begin{figure}[t]
    \begin{center}
    \includegraphics[width=0.8\columnwidth]{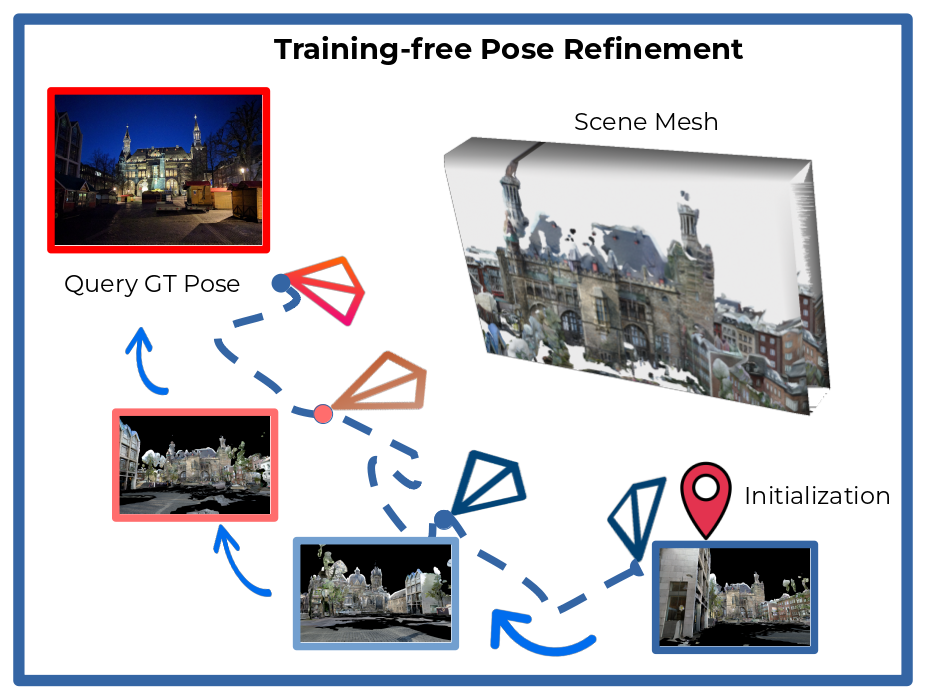}
    \end{center}
    \vspace{-0.2cm}
    \caption{\textbf{{\ours}} localizes images with a \textit{render\&compare} strategy. Given a starting hypothesis, a particle filter is used to perturb it and sample new candidates, which are rendered, and compared to the query using generic pre-trained features.
    }
    \vspace{-0.3cm}
    \label{fig:teaser}
\end{figure}
A feature-agnostic alternative to the previous map representations are meshes~\cite{Panek2022ECCV,panek2023visual,ventura2023p1ac}, 
as they support different tasks in the ecosystem of pose estimation, such as SLAM~\cite{newcombe2011denslam,sucar2021imap,bloesh2019meshslam,zhu2022nice}, tracking~\cite{labbe2022megapose}, path planning \cite{hudson2021heterogeneous}, and relocalization~\cite{arnold2022map,ventura2023p1ac} while providing the 3D information necessary for visual localization. Such models are easily obtained \cite{brejcha2020landscape,mueller2019image,kazhdan2013poisson}, and are rendered rather efficiently (\eg, 1 ms or less) even for large, textured models \cite{Panek2022ECCV}, relying on mature graphics primitives. \\
A different approach to visual localization is to refine an initial pose estimate. This strategy can either be applied to refine a pose estimate obtained from 2D-3D matches, or to obtain a more accurate pose starting from an initial hypothesis provided by image retrieval \cite{sarlin2018leveraging,humenberger2022investigating}. As such, these methods are to a large degree complementary to the matching-based methods described above. These approaches typically follow a \textit{render\&compare} framework~\cite{labbe2022megapose,li2018deepim}: in each iteration, a rendering (either an image \cite{labbe2022megapose,yen2021inerf,zhang2021reference} or the projection of a sparse set of features \cite{sarlin21pixloc,germain2023fqn,von2020lm}) of the scene obtained from the current pose estimate is compared to the actual image. Based on this comparison, an update is computed for the pose in order to better align the query with the scene representation. Existing approaches learn specific features for this task \cite{sarlin21pixloc,germain2023fqn,moreau2023crossfire,chen2023refinement}, potentially optimized together with the scene representation \cite{chen2022dfnet,moreau2022lens,liu2023nerf}. \\
We argue that in a \textit{render\&compare} framework, the main requirement is being able to evaluate the visual similarity of a synthetic view versus a real image. It has been shown repeatedly that generic deep features are a reliable estimator of this measure \cite{gatys2016style,zhang2018pips,kim2017visual}, and that this property of dense features makes them suitable to re-rank poses \cite{taira2018inloc}.
This is in contrast with the aforementioned refinement approaches, which rely on sparse features that need to be optimized for the task, and it leads us to the research question of whether it is truly essential to train specialized features for localization, or if analogous results can be attained exploiting the properties of dense features from generally available, off-the-shelf architectures. \\
Opting out of feature optimization also removes the need to articulate a differentiable feature-to-pose pipeline, required to compute gradients. Instead, to refine the pose, we adopt a simple particle filter-based optimizer \cite{thrun2005probabilistic,kwon2007filter} that efficiently explores the hypothesis space \cite{Soatto2000filterLie}. 
Despite its simplicity, our \textbf{\ours} approach outperforms modern pose regressors \cite{moreau2022lens,chen2022dfnet} and is comparable or better than refinement pipelines based on implicit fields \cite{moreau2023crossfire,germain2023fqn,chen2023refinement}, even though both these families of methods are optimized per-scene. Unlike them, our method also scales to large scenes. \\
While matching-based methods still hold the state-of-the-art, our method brings complementary strengths, in that it can be used to improve the performance of matching-based approaches as a post- or pre-processing step. 
We demonstrate these strengths through extensive experiments, both indoor and outdoor, as well as large scale scenarios, providing evidence that it is possible to construct a pose refiner that generalizes across different domains and representations, without the need for specialized training.

\textbf{Contributions}:
\begin{itemize}
    \item A simple yet powerful particle-filter based optimization which can be applied to different scene representations and scoring functions
    \item We provide an analysis on the effectiveness of general, pre-trained features at different layers of deep networks as a robust cost function
    \item We show a versatile pose-refinement approach which does not entail per-scene training or fine-tuning, that can be used either standalone, or to obtain a better pose prior, or to refine previous pose estimates
    \item The code, which allows to experiment with different backbones, scoring functions and scene representations, is available at {\small{\url{https://github.com/ga1i13o/mcloc_poseref}}}
\end{itemize}

\section{Related works}
\label{sec:rw}

\myparagraph{Visual Localization}
Visual Localization aims at estimating the camera pose of a given query within a known environment.
A popular strategy is to rely on sparse 3D models, obtained from SfM \cite{schoenberger2016sfm}, to represent the scene. These point clouds associate to each 3D location features triangulated from the available database. For inference, local features are used to find matches between a query and the 3D model \cite{sattler2015hyperpoints,sattler2016efficient,li2012worldwide,schonberger2018semantic,sarlin2019coarse,sarlin2020superglue,brejcha2020landscape,taira2018inloc,taira2019right}. 
Once 2D-3D matches are obtained, the query pose can be estimated via a PnP solver \cite{Larsson2019ICCV}.
To avoid matching against the entire database, it is common to apply a hierarchical approach \cite{sarlin2018leveraging,humenberger2022investigating,humenberger2020robust}, where a network for Place Recognition \cite{pion3DV2020, berton2022benchmark} selects database images with potentially covisible regions
\cite{arandjelovic2016netvlad,trivigno2023ICCV,berton2022cosPlace}. 
There also exist methods that replace the SfM model in favor of more versatile dense representation, either point clouds from Multi-View stereo or LIDAR \cite{schoenberger2016mvs, shan2014mvs, taira2018inloc} a mesh \cite{Panek2022ECCV,brejcha2020landscape,zhang2021reference}, or NeRFs \cite{mildenhall2020nerf,yen2021inerf,liu2023nerf}.
In this work we show that by relying on a renderable representation of the scene, it is possible to align the query pose by comparing features pixelwise, without resorting to exhaustive feature matching.
While matching-based methods retain state-of-the-art performances, our method has complementary strengths: it can be effortlessly applied either to refine initial poses or final estimates, improving matching results while adding little computational overhead.
\begin{figure*}[t]
    \begin{center}
    \includegraphics[width=0.9\linewidth]{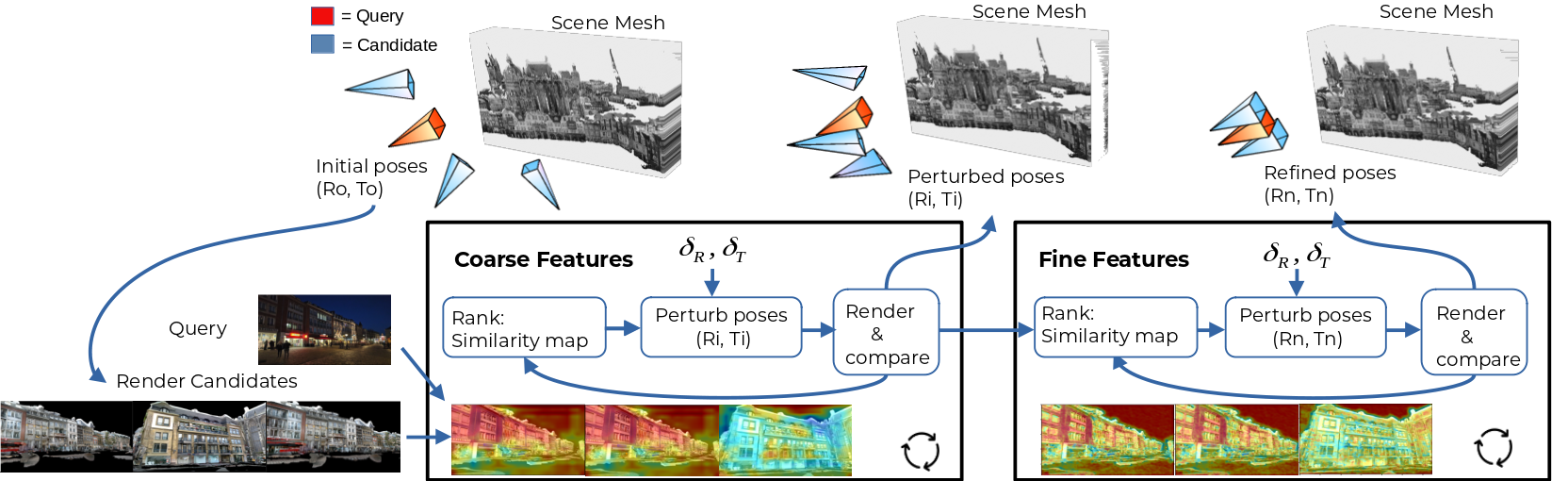}
    \end{center}
    \vspace{-0.25cm}
    \caption{\textbf{Architecture of {\ours}}. It exemplifies our iterative pose refinement. Given an initial pose estimate, we perturb it and render new candidates. Candidates are ranked based on dense, pixelwise feature similarity. As optimization progresses, we exploit the hierarchical properties of deep features by switching to shallower features, which are better for fine-grained comparison. 
    }
    \vspace{-0.2cm}
    \label{fig:architecture}
\end{figure*}

\myparagraph{Implicit representations for Visual Localization}
Both sparse and dense models store explicit information about the geometry of a scene. On the other hand, a parallel line of research has focused on implicit representations, that embed information about the scene into the weights of a neural network. Traditionally this was done either by training models to regress the camera pose \cite{kendall2017geometric,moreau2022lens,shavit2021ijcv,ding2019camnet}, or with Scene Coordinate Regressors, which encode for each image patch the corresponding 3D points\cite{brachmann2017dsac,brachmann2018learning,cavallari2019let,cavallari2020realtime}. 
More recently, neural radiance fields gained popularity \cite{mildenhall2020nerf,barron2021mip,muller2022instant}. These methods map each point of the scene to a view-dependent color and density values, through a MLP-based network.
These representations can be exploited for localization by embedding features in the implicit representation \cite{germain2023fqn,moreau2023crossfire,liu2023nerf} or by inverting the neural field \cite{yen2021inerf,lin2023parallel}.
Implicit representations have also been used as data augmentation to generate samples to train pose regressors \cite{moreau2022lens,chen2022dfnet}.

\myparagraph{Pose refinement and image alignment}
Pose refinement is a relevant area of research, in which the main idea is to iteratively refine the pose estimate by minimizing an objective function. In this family, a longstanding approach is represented by Direct Alignment methods, which minimize differences in pixel intensities when projecting the scene into the current estimate \cite{LK20years,engel2017direct}, using gradient-based optimizers such as Levenberg-Marquardt \cite{levenberg1944method,marquardt1963algorithm} or Gauss-Newton methods. These approaches are popular in SLAM scenarios \cite{schops2019bad,alismail2017photometric}, and typically rely on photometric error, hardly robust to appearance variations. They have been applied on learned features as well \cite{von2020gn,von2020lm,sarlin21pixloc}.
Indirect methods define geometric correspondences in order to minimize the reprojection error \cite{pietrantoni2023CVPR}.
\TSr{Actually, direct methods, at least in SLAM optimize a photometric or featuremetric error. Indirect methods use correspondences and minimize reprojection errors.}Among direct methods, a notable example is PixLoc \cite{sarlin21pixloc}, which trains features end-to-end from pixels to camera pose. Inference is performed via feature-metric alignment relying on the SfM model. 
Lately, pose refinement methods based on implicit representations have gained popularity. \cite{yen2021inerf} learns a radiance field that is used to render candidates, for which a photometric errors is computed and backpropagated.
Alternatively, implicit representations can be used to model a feature field, rather than appearance. Within these methods, FQN \cite{germain2023fqn} relies on reprojection error, whereas \cite{moreau2023crossfire,chen2023refinement} match the rendered features and then invert the descriptor field by backpropagating errors.
These approaches require to train features per-scene, and are only applicable for small scenes, given the limited scalability of implicit models. Our method relies on general, pre-trained features, which work on any dataset. Moreover, being agnostic to the scene representation, it can scale to arbitrarily large scenes where a mesh can be easily obtained \cite{Panek2022ECCV}. Our findings also relate to \cite{zhang2018pips}, that showed how deep learning models are surprisingly good at evaluating image similarities, outperforming all ``handcrafted" metrics. We extend this analysis beyond perceptual similarity and show how generic features can discern among fine-grained pose discrepancies.

\myparagraph{Localization with Particle filters}
Our work is not the first to employ a particle filter for localization \cite{poglitsch2015filter}. A similar optimization technique to ours is adopted in \cite{lin2023parallel,maggio2023loc}, although both these methods require a radiance field of the scene, thus sharing the limitations listed above, and rely on the less-than-robust pixel error. \\
Such approaches are popular also in mobile robotics, where they have been used for localization\cite{karkus2018particle,Fox2001,inam2009selfpart} and visual-tracking \cite{choi2012beams}. They have also been used in remote-sensing to localize against satellite images \cite{hu2020image}. Theoretical properties of particle filters have been studied in \cite{Soatto2000filterLie,kwon2007filter,busam2017camera,Kwon2008filter}.

\TSr{There is probably more work on localization with particle filters. I know that in the context of a moving platform, it has been used a lot in robotics, \eg, see \url{https://users.umiacs.umd.edu/~fer/cmsc828/classes/fox.mcmc-book.pdf}, \url{https://scholar.google.com/citations?view_op=view_citation&hl=en&user=ZxXBaswAAAAJ&citation_for_view=ZxXBaswAAAAJ:u5HHmVD_uO8C} and \url{https://arxiv.org/pdf/1903.00159.pdf}}

%%%%%%%%%%%%%%%%%%%%%%%%%%%%%
\section{\ours}
\label{sec:method}

\begin{figure}[t]
    \begin{center}
    \includegraphics[width=\columnwidth]{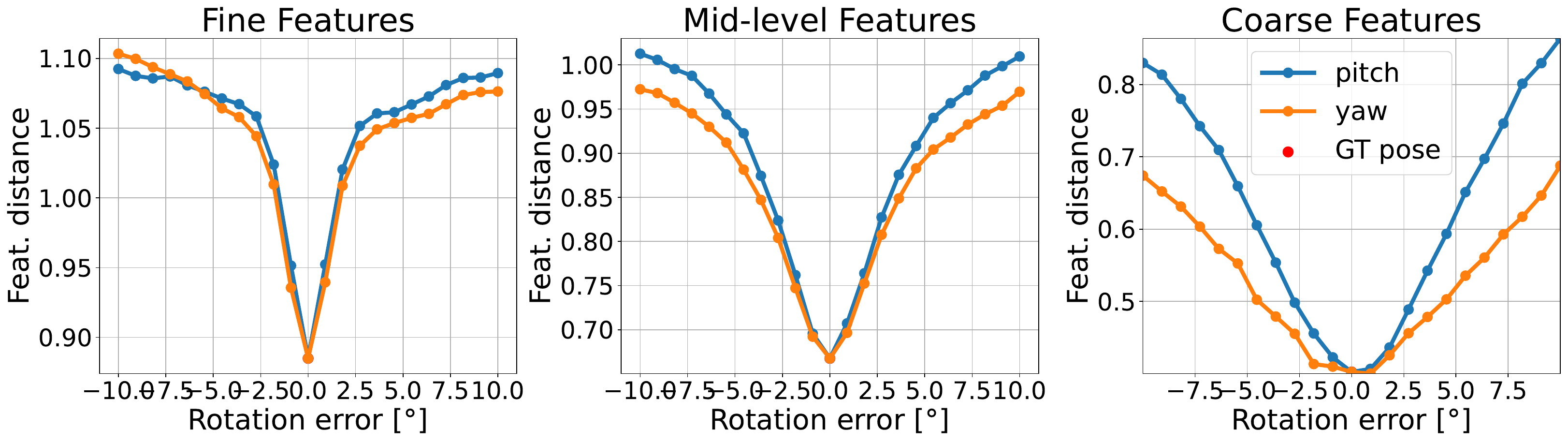}
    \includegraphics[width=\columnwidth]{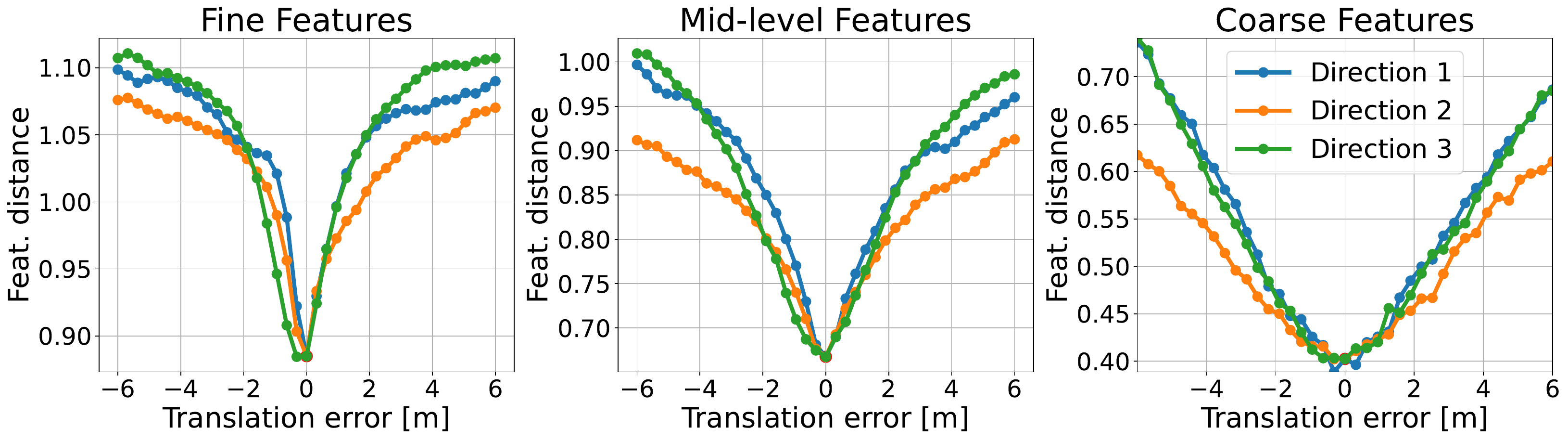}
    \end{center}
    \vspace{-0.4cm}
    \caption{\textbf{Convergence Basin in Optimization Space at Multiple Scales}. We perturb rotation and translation for a query from Aachen and compute the dense, pixelwise feature distance at different depths. 
    First row: rotating along yaw and pitch axis. Second row: moving away from the GT along 3 random directions.
    }
    \vspace{-0.3cm}
    \label{fig:optim_space}
\end{figure}

\myparagraph{Overview} {\ours} localizes a query image within a \textit{render-and-compare} framework, powered by Monte Carlo simulation. Localization is performed through iterative pose-refinement, as exemplified in \cref{fig:architecture}.
Given a query, and an initial hypothesis of its pose (which can be obtained in several different ways), we perturb it with some noise and rely on a renderable representation of the scene to generate the corresponding views. A simple, generic feature extractor is used as a cost function to evaluate which candidates are more similar to the query. The pose estimates are modeled and perturbed with a particle filter \cite{thrun2005probabilistic}, which serves as a stochastic optimizer. \\
Our method is agnostic to the scene representation used to render candidates, and we show that a general purpose feature extractor is suited to evaluate pose alignment, with no need for fine-tuning or per-scene training. 

\myparagraph{Motivation}
This work aims at answering the following research question: \textit{do we really need to train specialized descriptors or can generic features be used for localization?}. This question is rooted in the observation that activations of deep network are extremely reliable estimators of \textit{perceptual similarity}~\cite{gatys2016style,zhang2018pips}, being also robust to domain changes, blur and distortion \cite{kim2017visual}. 
Intuitively, perceptual similarity seems a promising metric for measuring pose similarity via a \textit{render \& compare} approach. We show that this property, coupled with the natural spatial structure of feature maps, yields a simple and effective tool to  measure pose discrepancies using perceptual similarity as a way to measure pose similarity. To this end, we integrate a perceptual metric into a particle-filter-based optimizer \cite{thrun2005probabilistic,choi2012beams}, that is used to generate new pose hypothesis to be then rendered \& compared.
%and obtain a method that generalizes across different datasets, scene representations and feature extractors, with no need of any per-scene tuning.

\myparagraph{Problem setting}
Our objective is, for a given query image $I_q$, to estimate its 6-DoF pose. 
Following \cite{kendall2015posenet,humenberger2022investigating}, we parametrize the pose as  $T_q = (\mathbf{c}, \mathbf{q})$, where $\mathbf{c} \in \mathbb{R}^3$  represents the camera center and $q \in \mathbb{R}^4$ is a unit quaternion.  
Quaternion-based parametrizations provide a framework for manipulating rotations which is numerically stable, compact and avoids gimbal lock~\cite{lepetit2005quat}. This formulation decouples translation and rotations updates, lying on the manifold of $\mathrm{SO}(3) \times \mathrm{T}(3)$ \cite{busam2017camera,lin2023parallel}. \\
We cast the problem as the following optimization:
\begin{equation}
\label{eq:argmin}
    \hat{T}_q = \underset{T \in \mathrm{SO}(3) \times \mathrm{T}(3)}{\text{argmin}} \ \mathcal{L_{\mathcal{F}_\theta}}(T | I_q, I_T)
\end{equation}
where $I_q, I_T$ are the query image and the rendered candidate with pose $T$, $\mathcal{F_{\theta}}$ is a feature extractor, and the loss function is the distance between query and rendered candidate in feature space, \ie, $\mathcal{L_{\mathcal{F}_\theta}}(T | I_q, I_T) = || \mathcal{F}_\theta (I_q) - \mathcal{F}_\theta (I_T)  ||_2$. 
We optimize this loss via a particle filter-based approach.

\subsection{Pose alignment with Pre-trained features}
\label{sec:features}

To evaluate the loss in \cref{eq:argmin} associated with a candidate pose \wrt the query, we forward both through an off-the-shelf CNN. More details on the specific architecture will be discussed later on. 
We obtain a hierarchy of feature volumes, $F_l \in \mathbb{R}^{C_l \times H_l \times W_l}$, for each level $l \in \{ 1..L\}$. These feature pyramids have decreasing resolution, and encode increasingly richer semantic clues as the receptive field of each neuron grows. It has been demonstrated as an \textit{emergent property} \cite{gao2017sim,zhang2018pips} that such hierarchies of features can measure perceptual similarities at different conceptual levels \cite{amirshahi2016image}; to the best of our knowledge no previous works leverage this property of dense feature maps to assess \textit{pose similarity} in a pose refinement algorithm. \\
We employ a simple scoring function that exploits this property;
at a given step $s$ of our optimization, we choose level $l(s)$ to compute the score of a candidate $I_T$ against query $I_q$ as follows:
\begin{equation}
\label{eq:score}
\begin{aligned}
    S(h, w | l) = \Bigg\| \frac{F_l^{h, w}(I_q)}{||F_l^{h, w}(I_q)||_2} &- \frac{F_l^{h, w}(I_T)}{||F_l^{h, w}(I_T)||_2} \Bigg\|^2  \\
    \mathcal{L_{\mathcal{F}_\theta}}(T | I_q, I_T, l) = &\frac{1}{h_l  w_l}\sum_{h,w}{S(h, w | l)}
\end{aligned}
\end{equation}
where $F_l^{h,w} \in \mathbb{R}^{C_l}$. 
In practice, we compare pixelwise dense, normalized descriptors, obtaining a spatial similarity map $S \in \mathbb{R}^{H_l \times W_l}$, which is then averaged. \\
In the early stages of the optimization, we need to deal with large baselines as initial hypothesis might be far off from the ground truth. Thus, to increase the convergence basin, we adopt a hierarchical \textit{Coarse-to-Fine} approach. Initially we rely on deeper features: their receptive fields are larger, hence even if the poses deviate significantly, there is a higher chance for two receptive fields of a pixel position to overlap, providing a meaningful signal. Moreover, since their features are semantically richer, they are more prone to ignore low-level details, transient objects and artifacts introduced by the scene representation. This allows to handle misalignment to a certain degree. 
As the optimization converges towards more accurate poses, the focus becomes discerning fine-grained details and small orientation displacements. At this stage, shallower features are more suited, as we can exploit their smaller receptive fields and  higher spatial resolution.
We discover that pre-trained features, across different architectures and training methods, are \textit{unreasonably effective}, as deemed in \cite{zhang2018pips}, meaning that they present a nicely shaped convex basin around each pose. We experimentally demonstrate this finding in \cref{fig:optim_space}, which also illustrates how moving up the hierarchy we can control the width of the basin, and that shallower features are able to precisely discern among even the finer differences. 

\begin{figure}[t]
    \begin{center}
    \includegraphics[width=\columnwidth]{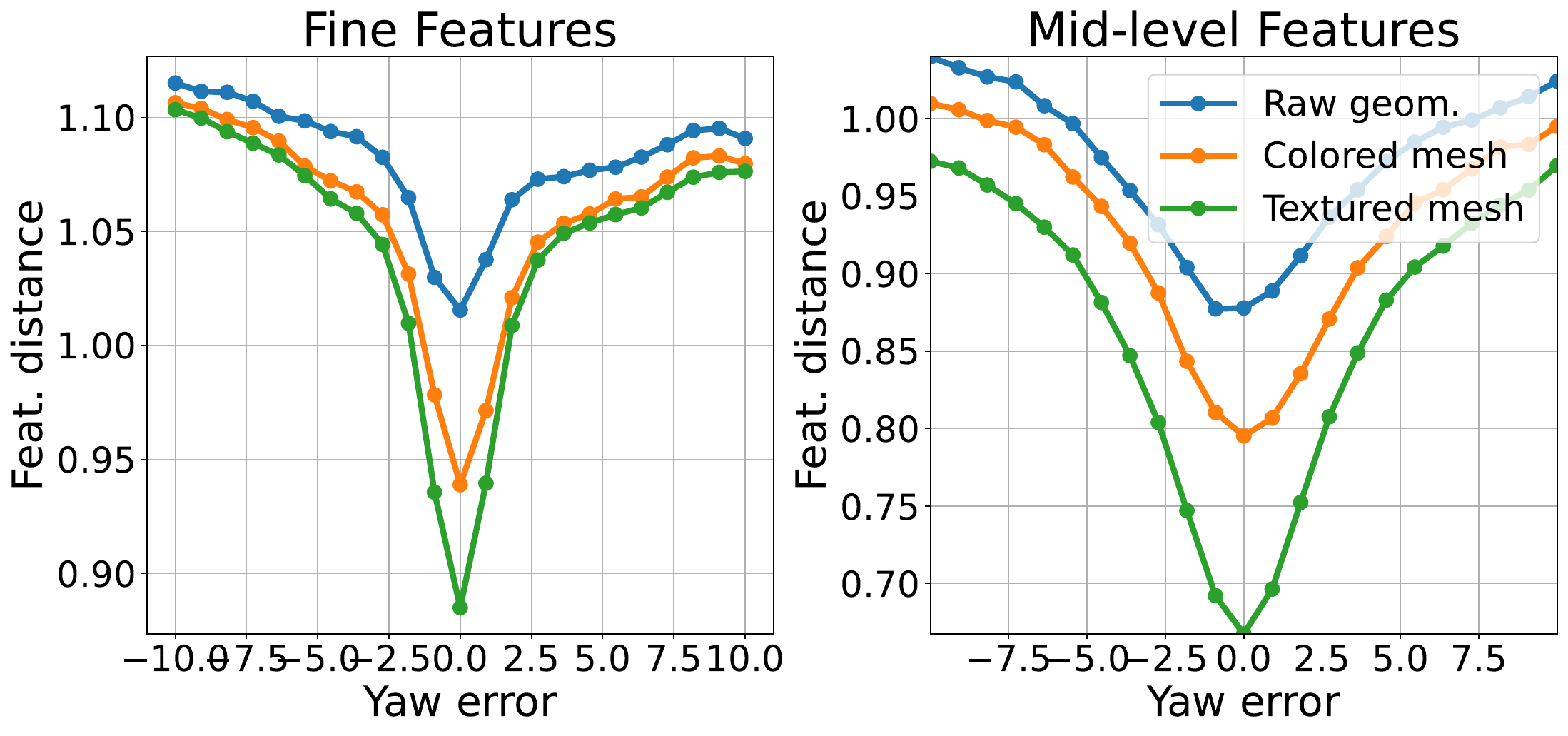}
    \includegraphics[width=\columnwidth]{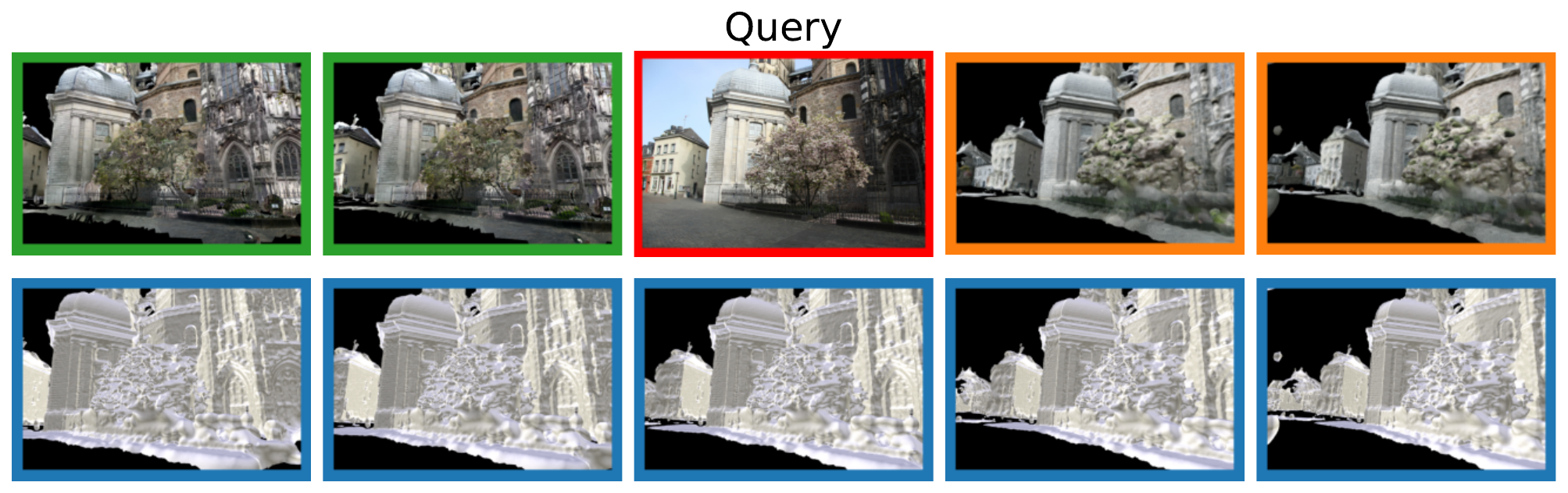}
    \end{center}
    \vspace{-0.4cm}
    \caption{\textbf{Robustness of the Convergence Basin to the Rendering Domain}. We render images rotating along the yaw axis, using different meshes:Textured, Colored and Raw Geometry, and evaluate the feature distance at different depths. The domain shift affects absolute values but not the basin shape.
    }
    \vspace{-0.3cm}
    \label{fig:domain}
\end{figure}

\subsection{Particle filter optimization}
\label{sec:part_filt}
\myparagraph{Overview}
Particle filters are a set of Monte Carlo methods that estimate the state of a system based on observations and dynamics of the system \cite{thrun2005probabilistic,kwon2007filter}. 
Such algorithms can approximate a wide range of distributions, and are computationally efficient since they focus on regions of the state space with high likelihood \cite{Fox2001}. The application of particle filters for visual localization is not novel, and their effectiveness has been demonstrated in \cite{poglitsch2015filter,lin2023parallel,karkus2018particle,maggio2023loc}. \\
The basic idea is that, given a starting hypothesis, by perturbing the initial state we can obtain new state hypothesis, evaluate the cost function and refine the estimate iteratively. In our case, the state variable is the camera pose $T$ and the scoring function to evaluate an hypothesis is the \textit{perceptual similarity}. 
Specifically, at each step the particle filter models the posterior distribution $p( \textbf{T}_q | \textbf{Z}_i)$ of the query pose $\textbf{T}_q$, with a set of particles $\textbf{Z}_i = \{ (T_i^1, \pi_i^1), .. (T_i^n, \pi_i^n) \}$. Particles have a weight $\pi_i^n$ that represents their likelihood, estimated via \cref{eq:argmin}. Since the particles states $T_i^1, \ldots, T_i^n$ are parameterized on the $\mathrm{SO}(3) \times \mathrm{T}(3)$ manifold, we can perturb them using their Lie algebra, as it was proven by \cite{Soatto2000filterLie, Kwon2008filter, choi2012beams} that particle filtering on Lie groups is coordinate-invariant, \ie, same perturbation on different states (poses) results in the same motion.

\myparagraph{Our approach}
In general, the loss function in \cref{eq:argmin} is not convex over the 7D optimization space, and the convergence basin is highly affected by initialization. Thus the main challenges are: exploring efficiently the otherwise large hypothesis space, and increasing the convergence basin.
To address the former, we rely on multi-hypothesis tracking \cite{choi2012beams} of multiple \textit{beams}. With \textit{beam} we denote a set of particles that evolves and is optimized independently from other \textit{beams}. This is equivalent to having separate optimization threads. It allows to explore in breadth the state space \cite{douc2005comparison}, and if some threads get stuck in local minima, it does not affect the others.
The beams are optimized in parallel, to augment the probability that some of them will move in the right direction. Additionally, to minimize the cost of these initial steps, candidates can be rendered at low resolution ($256 \times 340$), as fine-grained details are not needed at this stage. Every $N_0$ iterations, a \textit{resampling step} is performed. The best candidates are pooled among all the beams, and each of them is used to initialize a new beam that is optimized independently again for $N_0$ steps. In this way we avoid pursuing unpromising hypothesis as beams that do not converge to good poses are halted. \\
To enhance convergence probability, the \textit{Coarse-to-Fine} strategy discussed in \cref{sec:features} is adopted. Following this reasoning, after $N_1$  \textit{resampling steps}, we switch to shallower feature maps in our feature extractor. 
Additionally, over the iterations image resolution is gradually increased, while the number of beams and particles in each beam is decreased.
This strategy enables to keep computational cost low, while balancing the need to explore in breadth the sample space in the beginning, and to have fine-grained comparisons as we refine further the pose. 
More details on these hyperparameters are given in \cref{sec:implement}, and the pseudo-code of the algorithm is provided in the Supplementary. The code will be publicly released upon acceptance.

\subsection{Adapting to different domains}
\label{sec:domains}
Given that our framework entails comparing query images against rendered candidates, it raises the question of whether we might need an adaptation technique to bridge the gap between domains. Recently, this issue was addressed in \cite{Panek2022ECCV,zhang2021reference}, showing that matching performances are not hindered by the rendering domain.
We extend their analysis since in our setup we compare dense feature maps, which is different than matching local descriptors. \\
We test different rendering domains (textured, colored, raw geometry) and find that these shifts indeed cause a discrepancy in the extracted features, meaning that the distance between a query and the rendered ground truth pose will not be 0. Nevertheless, we are not interested in absolute values, as the only requirement for our optimization to converge is that relative differences in pose are reflected by relative changes in similarity. \Cref{fig:domain} exemplifies this effect. \\
We show that in practice this assumption holds, and different domains only affect absolute values, preserving relative differences, since the rendered images domain is uniform.
This finding highlights an advantage of our formulation, being agnostic to the scene representation.

\begin{figure}[t]
    \begin{center}
    \includegraphics[width=0.8\columnwidth]{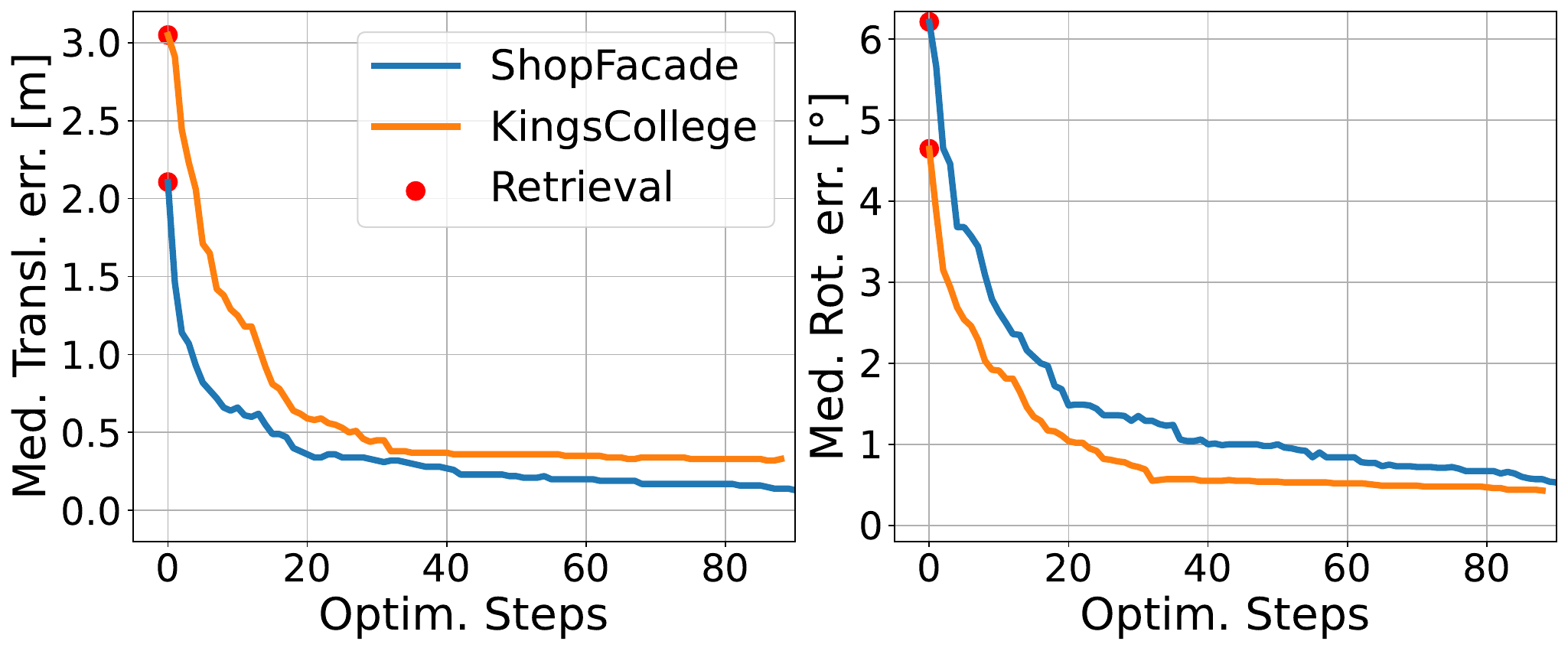}
    \end{center}
    \vspace{-0.4cm}
    \caption{\textbf{Optimization trajectory}. Behavior of median errors over the iterations for 2 scenes from Cambridge Landmarks.
    }
    \vspace{-0.4cm}
    \label{fig:trajectory}
\end{figure}
\section{Experiments}
\label{sec:exp}
\myparagraph{Datasets}
\label{sec:datasets}We evaluate our pose refinement approach on multiple datasets. Aachen Day-Night v1.1 \cite{zhang2021reference, sattler2012image, sattler2018benchmarking}  is a common benchmark for large-scale localization \cite{sarlin21pixloc,sarlin2019coarse}; it contains  6,697 day-time database images and 1,015 queries, collected by handheld devices. Beyond the large area that it covers, it contains night queries and strong viewpoint changes between the database and query images. We also evaluate on smaller datasets widely used in the literature, namely Cambridge Landmarks \cite{kendall2015posenet} and 7scenes \cite{shotton2013scene}. They contain respectively 5 outdoor and 7 indoor scenes. 
In both datasets query sequences are captured along different trajectories \wrt the available database. 
Following common practices, we report for Aachen the recall at thresholds $(25\text{cm}, 2\degree)$, $(50\text{cm}, 5\degree)$, and $(5\text{m}, 10\degree)$ \cite{sattler2018benchmarking,sarlin2019coarse}, while for the remaining datasets we evaluate median translation (m) and rotation ($\degree$)  errors \cite{moreau2023crossfire,liu2023nerf,moreau2022lens}.

\begin{table}[th]
    \centering
    \begin{center}
    \begin{adjustbox}{width=0.9\columnwidth}
\begin{tabular}{ll|ccccccc}
    \toprule
    \multirow{2}{*}{Coarse Features} & \multirow{2}{*}{Fine Features} & \multirow{2}{*}{ShopFacade} & \multirow{2}{*}{OldHospital}\\
    &&& \\
    \midrule
    \textit{ResNet-18} & & \\
    ~CosPlace \cite{berton2022cosPlace} &  ImageNet &
    12 / 0.45 & 39 / 0.73 \\
    ~ImageNet  & ImageNet & 12 / 0.55 & 46 / 0.80 \\
    ~SimCLR \cite{chen2020simple} & SimCLR \cite{chen2020simple} & 18 / 0.62 & 50 / 0.83 \\
    \midrule
    ALIKED \cite{zhao2023aliked}   &ALIKED \cite{zhao2023aliked}   & 17 / 0.64 & 49 / 0.84 \\
    AlexNet \cite{krizhevsky2012anet} &AlexNet \cite{krizhevsky2012anet} & 15 / 0.74 &  53 / 0.88 \\
    \bottomrule
\end{tabular}
\end{adjustbox}
    \end{center}
    \vspace{-0.3cm}
    \caption{\textbf{Ablation on feature extractors}, shows that the property of dense features being robust estimators of \textit{visual similarity} holds across architectures and training protocols. Errors in $cm, \degree$.
    }
    \vspace{-0.3cm}
    \label{tab:abl_model}
\end{table}

\subsection{Implementation details}
\label{sec:implement}
For our main experiments, we adopt a lightweight ResNet-18 \cite{He2016resnet} trained for Place Recognition in \cite{berton2022cosPlace}. This network was fine-tuned from \textit{conv3}, freezing earlier layers. Thus when in our optimization we switch to \textit{conv2}, we are actually using ImageNet features.
We find that this network slightly improves results over vanilla ImageNet (see \cref{tab:abl_model}). We hypothesize that being trained for Place Recognition, deeper layers have learnt to focus on buildings and landmarks, ignoring transient objects, which is useful also for localization. After $N_1$ steps, we switch to \textit{conv2} features, pre-trained only on ImageNet; finally, the last refinement steps (after $N_2$ iterations), are performed with \textit{conv1}. The choice of $N_1, N_2$ is not critical to achieve good results; what matters is that initial steps are carried out with coarser features, and the very last with finer ones, as \textit{conv1} features present a narrower converge basin. In the Supplementary we report experiments to demonstrate robustness to these hyperparameters, and a convergence analysis. 
Their values also depend on the use-case: when using our algorithm as post-processing, there is no need to start from Coarse features; viceversa when acting as pre-processing we do not use the shallower features. For the standalone experiments starting from retrieval poses, we set $N_1$ to 30 for all datasets, and $N_2$ to $N_t - 10$, \ie the last 10 steps are with \textit{conv1}. \\
In \cref{tab:abl_model}, we ablate the choice of network, showing that our method works with any off-the-shelf architecture.
For perturbing the camera center, we use Gaussian noise, with the only precaution that the standard deviation on the vertical axis is reduced to $1/10$ wrt the other directions. This comes from the prior knowledge that while we need to explore the scene in breadth, as initialization can be far off, height variations are typically limited to human size, thus we can avoid wasting resources exploring the vertical axis.
We perturb the rotation around a random axis, with uniform noise.
The magnitude of the noise is reduced linearly, and every $N_0=20$ steps it is reset. This \textit{start-and-stop} scheduling is akin to the CosineAnnealing strategy \cite{loshchilov2016sgdr}.

\myparagraph{Renderable scene representations}
\label{sec:scene_repr}
%We experiment with different kinds of scene representation.
The only requirement for our method is to have a renderable  model of the scene; \ie that allows to generate a view given any pose $(R, t) \in \mathrm{SE}(3)$. Recently, \cite{Panek2022ECCV,panek2023visual} highlighted the 
advantages of 3D meshes and how they can be obtained. Specifically, these advantages are flexibility of supporting different tasks, and efficient rendering, thanks to rendering pipelines being tailored to meshes for decades.
As an alternative to meshes, modern neural radiance fields \cite{mildenhall2020nerf,nerfstudio} offer photorealistic renderings, at the cost of being typically slower. Although several efforts greatly cut down on NeRF rendering times \cite{mueller2022instant,Reiser2021ICCV}, they remain at least 1 order of magnitude slower than mesh-based rendering.
In light of this, we experiment with Gaussian Splatting \cite{kerbl3Dgaussians}, which offers high-quality images and matches the speed of a mesh.
In our experiments we find that a low-detail, compressed mesh is sufficient to achieve good results. 
On the large scale scene of Aachen \cite{sattler2012image,sattler2018benchmarking,zhang2021reference}, we use the models provided by \cite{Panek2022ECCV}. Thanks to the mesh being compressed, and the low resolutions that we adopt, a textured image can be rendered efficiently in $500 \mu s$. \TSr{Do we want to say that we are using the meshes from \cite{Panek2022ECCV} here and reference that paper for the rendering times?}
For smaller scenes of Cambridge Landmarks~\cite{kendall2015posenet} and 7scenes~\cite{shotton2013scene}, starting from the point cloud of the dataset we optimize a set of 3D Gaussians following \cite{kerbl3Dgaussians}, which requires only $10 min$, and can then be rendered in $600-900 \mu s$, depending on the scene. Times were measured on a RTX 4090 GPU.

\begin{table}[t!]
    \begin{center}
    \begin{adjustbox}{width=0.95\columnwidth}
\begin{tabular}{llccccccccccccc}
    \toprule
    \multirow{2}{1cm}[-.4em]{Method}  & & \multicolumn{4}{c}{Cambridge Landmarks}\\
    \cmidrule(lr){3-6}
    &&  King's & Hospital & Shop & St. Mary's \\
    \midrule
    \textit{Retrieval} &&&& \\
    ~DenseVLAD~\cite{torii201524} & - & 2.8/5.7  & 4.0/7.1 & 1.1/7.6 & 2.3/8.0  \\
    ~CosPlace~\cite{berton2022cosPlace} & -& 3.1/4.4  & 4.5/6.7 & 2.1/6.2 & 3.2/7.2  \\
    \midrule
    \textit{SOTA}  &&&&& \\
    ~AS~\cite{sattler2016efficient}$^\dag$  & -& 0.13/0.22 & 0.20/0.36 & 0.04/0.21 & 0.08/0.25\\
    ~hloc~\cite{sarlin2019coarse}  & TL & 0.12/0.20 & 0.15/0.30 & 0.04/0.20 & 0.07/0.21 \\
    ~DSAC* \cite{brachmann2020dsacstar} &TS& 0.15 / 0.3 & 0.21 / 0.4 & 0.05 / 0.3 & 0.13 / 0.4  \\
    ~HACNet \cite{li2020hierarchical} &TS& 0.18 / 0.3 & 0.19 / 0.3 & 0.06 / 0.3 & 0.09 / 0.3  \\
    ~PixLoc \cite{sarlin21pixloc}&TL&  0.14/0.24 & 0.16/0.32 & 0.05/0.23 & 0.10/0.34\\
    \midrule
    \textit{Pose Regressors} &&&&& \\
    ~MS-Transformer \cite{shavit2021ijcv}& TS & 0.83 / 1.47 & 1.81 / 2.39 & 0.86 / 3.07 & 1.62 / 3.99 \\
    ~DFNet \cite{chen2022dfnet}& TS & 0.73 / 2.37 & 2 / 2.98 & 0.67 / 2.21 & 1.37 / 4.03 \\
    ~LENS \cite{moreau2022lens}& TS & 0.33 / 0.5 & 0.44 / 0.9 & 0.25 / 1.6 & 0.53 / 1.6 \\
    \midrule
    \textit{Pose Refiners} &&&&& \\
    ~FQN \cite{germain2023fqn} & TS & \textbf{0.28} / \textbf{0.4} & 0.54 / 0.8  & 0.13 / 0.6 & 0.58 / 2.0 \\
    ~CROSSFIRE \cite{moreau2023crossfire} & TS & 0.47 / 0.7 & \underline{0.43} / \textbf{0.7} & 0.2 / 1.2 & 0.39 / 1.4 \\
    ~NeFeS (DFNet) \cite{chen2023refinement} & TS &  0.37 / 0.62 & 0.55 / 0.9 & \underline{0.14} / \underline{0.47} & \underline{0.32} / \underline{0.99} \\
    ~\textbf{\ours~(ours)} & - & \underline{0.31} / \underline{0.42} & \textbf{0.39} / \underline{0.73} & \textbf{0.12} / \textbf{0.45} & \textbf{0.26} / \textbf{0.88} \\ 
    \bottomrule
\end{tabular}
\end{adjustbox}
    \end{center}
    \vspace{-0.2cm}
    \caption{\textbf{Results on the Cambridge Landmarks dataset.} We show that our simple approach outperforms methods that train per-scene descriptors. TM marks methods trained for feature matching, TL trained for localization, TS trained per scene.}
    \label{tab:small-loc}
    %\vspace{-0.1cm}
\end{table}

\begin{table}[t!]
    \centering
    \begin{adjustbox}{width=0.8\columnwidth}
\begin{tabular}{lcccccc}
    \toprule
    \multirow{2}{*}{Method} & \multicolumn{2}{c}{Aachen Day-Night v1.1}  \\
    \cmidrule(lr){2-3}
    & Day & Night \\
    \midrule
    %\multirow{2}{*}{\begin{sideways}IR\end{sideways}}
    \textit{Retrieval} && \\
    ~NetVLAD~\cite{arandjelovic2016netvlad} &
    0.0 / 0.2 / 18.9 & 0.0 / 0.0 / 14.3 \\
    ~CosPlace~\cite{berton2022cosPlace} &
    0.0 / 0.4 / 27.1 & 0.0 / 0.0 / 24.1 \\
    \midrule
    %\multirow{2}{*}{\begin{sideways}IT\end{sideways}}
    \textit{Pose Refiners} && \\
    ~Pixloc~\cite{sarlin21pixloc} &
    \underline{63.2} / 67.8 / 75.5 & 38.7 / 47.1 / 60.7 \\
    ~\textbf{\ours~(ours)} &
    55.8 / \underline{73.3} / \underline{89.7} & \underline{42.4} / \underline{66.5} / \underline{86.9} \\
    \midrule
    %\multirow{3}{*}{\begin{sideways}FM\end{sideways}}
    \textit{Matching based} && \\
    ~AS \cite{sattler2016efficient} & 85.3 / 92.2 / 97.9 & 39.8 / 49.0 / 64.3 \\
    ~hloc \cite{sarlin2019coarse} &
    87.4 / \textbf{95.0} / 98.1 & 71.7 / \textbf{88.5} / \textbf{97.9} \\
    ~ + PixLoc refine &
    86.2 / 94.9 / 98.1 & 70.8 / \textbf{88.5} / \textbf{97.9} \\
    ~ + \textbf{(ours)} refine &
    \textbf{87.9} / 94.9 / \textbf{98.9} & \textbf{73.8} / \textbf{88.5} / \textbf{97.9} \\
    \bottomrule
\end{tabular}
\end{adjustbox}
    \caption{\textbf{Large scale Visual localization on Aachen v1.1 dataset.} We show competitive results against PixLoc refinement, and how our method can be coupled with sota pipelines to improve results. 
    }
    \label{tab:large-loc}
    \vspace{-0.25cm}
\end{table}

\subsection{Experimental results}
\label{sec:results}
In this section, we perform an ablative study to support our choices and the motivations of the paper, together with visualizations to highlight salient aspects.
We then validate our results on against state-of-the-art (sota) matching methods, pose regressors and implicit features-based refiners.

\myparagraph{Ablation studies}
\label{sec:ablations}\cref{tab:abl_model} ablates different architectural choices. For all architectures, we extract dense features and use them as in \cref{eq:score}. While CosPlace+ImageNet achieves slightly superior results, our method works regardless of the architecture, training protocol and/or dataset. These results expand on the findings of the LPIPS paper \cite{zhang2018pips}, proving the \textit{unreasonable effectiveness} of generic features not only for perceptual similarity, but for pose similarity as well. 
In particular, \cref{tab:abl_model} shows that dense features, trained either on supervised or unsupervised objectives, for generic classification, place recognition or feature matching (ALIKED \cite{zhao2023aliked}), show the same property that is visualized in \cref{fig:optim_space}. 
That is, ability to estimate image alignment, with high precision in shallower layers, and with wider baselines in deeper features. 
These findings also align with the foundation behind transfer learning \cite{doersch2015unsupervised}, a cornerstone of modern computer vision, which can be summarized in very simple words as "a good feature is a good feature anywhere" \cite{zhang2018pips}. These networks, regardless of the architecture or the pre-training task, learned how to extract generic features, and we can exploit the natural spatial structure of the feature maps to discriminate pose variations.
In the Supplementary we also provide an ablation on different scoring functions, showing that a simple, dense 
pixelwise comparison is \textit{all you need}, against more elaborate formulations.

\myparagraph{Baselines}
To evaluate whether the generic features that we adopt are competitive, we benchmark against methods that train per-scene.
\begin{itemize}[noitemsep,topsep=1pt]
    \item \textbf{Pose Regressors}: These method train a network to directly predict the camera pose. We consider DFNet \cite{chen2022dfnet}, LENS \cite{moreau2022lens} and MS-Transformer \cite{shavit2021ijcv}
    \item \textbf{Pose Refiners}: These are the closest to our method. Among them, FQN \cite{germain2023fqn}, NeFeS \cite{chen2023refinement} and CROSSFIRE \cite{moreau2023crossfire} optimize per-scene descriptors in an implicit field and as such are limited to small scenes. PixLoc \cite{sarlin21pixloc} trains features specific for localization on MegaDepth \cite{li2018megadepth}. Their localization pipeline minimizes a feature-metric objective with first order methods (FQN/NeFeS/Crossfire) or second order optimizers (PixLoc), whereas we rely on a simple MonteCarlo algorithm
    \item \textbf{Matching based}: These methods represent the state-of-the-art. We show how our method can be coupled with them to further improve performances
\end{itemize}

\begin{table}[th]
    \centering
    \begin{center}
    \begin{adjustbox}{width=0.85\columnwidth}
\begin{tabular}{lcccccccc}
    \toprule
    \multirow{2}{*}{MeshLoc \cite{Panek2022ECCV} pipeline} & \multirow{2}{*}{\begin{tabular}[c]{@{}c@{}}Top K \\Matched\end{tabular}} & \multicolumn{3}{c}{\multirow{2}{*}{Aachen Night v1.1}} & \\
    &&& \\
    \midrule
    \textit{Textured Mesh} & & \\
    ~LoFTR \cite{sun2021loftr}  & 50 &
    73.3 / 89.0 / 95.8 \\
    ~LoFTR \cite{sun2021loftr}  & 20 &
    71.2 / 89.0 / 94.8 \\
    ~LoFTR \cite{sun2021loftr}  & 10 &
    70.7 / 86.4 / 94.8 \\
    ~\textbf{(ours)} + LoFTR \cite{sun2021loftr} & 20 &
    ~\textbf{74.3} / \textbf{91.1} / \textbf{99.5} \\
    ~\textbf{(ours)} + LoFTR \cite{sun2021loftr} & 10 &
    \underline{73.8} / \textbf{91.1} / \underline{99.1} \\
    \midrule
    \textit{Raw Geometry} & & \\
    ~P2P\cite{zhou2021patch2pix} + SG \cite{sarlin2020superglue} & 50 & 8.4 / 27.7 / 60.7 \\
    ~P2P\cite{zhou2021patch2pix} + SG \cite{sarlin2020superglue} & 10 & 6.8 / 20.4 / 52.4 \\
    ~\textbf{(ours)} + P2P\cite{zhou2021patch2pix} + SG \cite{sarlin2020superglue} & 1 & 16.8 / 37.7 / 66.0 \\
    \bottomrule
\end{tabular}
\end{adjustbox}
    \end{center}
    \vspace{-0.2cm}
    \caption{\textbf{Preprocessing on Aachen Night}: {\ours} can improve initial poses from retrieval, before a more expensive localizer.
    }
    \vspace{-0.3cm}
    \label{tab:preproc}
\end{table}

\myparagraph{Comparison with methods trained per-scene} 
In \cref{tab:small-loc}  we compare our method mainly against other refinement methods \cite{chen2023refinement,germain2023fqn,moreau2023crossfire} and pose regressors \cite{chen2022dfnet,moreau2022lens} on the Cambridge Landmarks benchmark \cite{kendall2015posenet}.
The main rationale of this set of experiments is to compare against approaches that train scene-specific descriptors and/or representations for localization.
While pose regressors surely achieve the faster inference time, they generally perform worse. 
Despite the absence of any kind of fine-tuning, we outperform all implicit feature-based refiners, except a small gap on King's College were FQN is slightly better. On these datasets, our optimization converges in 80 refinement steps, although satisfying results are achieved much earlier. \cref{fig:trajectory} shows the optimization trajectory on 2 distinct scenes. Overall, we conclude that to achieve satisfying results there is no need to fine-tune per-scene, or at all. Among methods that train per-scene, Scene Coordinate Regressors \cite{brachmann2020dsacstar,li2020hierarchical} perform best.

\begin{table}[th]
    \begin{center}
    \begin{adjustbox}{width=\columnwidth}
\setlength\tabcolsep{2pt}
\begin{tabular}{lcccccccc}
    \toprule
    \multirow{3}{*}{Method} & \multicolumn{7}{c}{\multirow{2}{*}{\begin{tabular}[c]{@{}c@{}}7 scenes: DSLAM ground truths\\ median error in ~(cm/\degree) $\downarrow$\end{tabular}}} \\
    &&&&&&& \\
    \cmidrule(lr){2-8}
    &Chess & Fire & Heads & Office & Pumpkin & Kitchen & Stairs  \\
    \midrule
    \textit{Retrieval} &&&&&&& \\
    ~DenseVLAD \cite{torii201524}& 21/12.5 & 33/13.8 & 15/14.9 & 28/11.2 & 31/11.3 & 30/12.3 & 25/15.8 \\
    ~CosPlace \cite{berton2022cosPlace} & 31 / 11.4 & 45 / 14.6  &  23 / 13.7 & 43 / 11.2 & 52 / 11.4 & 48 / 11.1 & 46 / 14.8 \\
    \midrule
    \textit{SOTA} &&&&&&& \\
    ~AS \cite{sattler2016efficient} &  3/0.87 & 2/1.01 & 1/0.82 & 4/1.15 & 7/1.69 & 5/1.72 & 4/1.01 \\
    ~DSAC \cite{brachmann2017dsac} &  2/1.10 & 2/1.24 & 1/1.82 & 3/1.15 & 4/1.34 & 4/1.68 & 3/1.16 \\
    ~HACNet \cite{li2020hierarchical} &  2/0.7 & 2/0.9 & 1/0.9 & 3/0.8 & 4/1.0 & 4/1.2 & 3/0.8 \\
    ~hloc \cite{sarlin2019coarse} & 2/0.85 & 2/0.94 & 1/0.75 & 3/0.92 & 5/1.30 & 4/1.40 & 5/1.47 \\
     \midrule
     \textit{Pose Regressors} &&&&&&& \\
     ~MS-Transf. \cite{shavit2021ijcv} & 11 / 4.7 & 24 / 9.6 & 14 / 12.2 & 17 / 5.66 & 18 / 4.4 & 17 / 6.0 & 17 / 5.9 \\
    ~DFNet\cite{chen2022dfnet} & 5 / 1.9 & 17 / 6.5 & 6 / 3.6 & 8 / 2.5 & 10 / 2.8 & 22 / 5.5 & 16 / 2.4 \\
    ~LENS \cite{moreau2022lens} & 3 / 1.3 & 10 / 3.7 & 7 / 5.8 & 7 / 1.9 & 8 / 2.2 & 9 / 2.2 & 14 / 3.6 \\
    \midrule
    \textit{Pose Refiners} &&&&&&& \\
    ~FQN-PnP \cite{germain2023fqn} & 4 / 1.3 & 10 / 3.0 & 4 / 2.4 & 10 / 3.0 & 9 / 2.4 & 16 / 4.4 & 140 / 34.7 \\
    ~CROSSFIRE \cite{moreau2023crossfire} & 1 / 0.4 & 5 / 1.9 & 3 / 2.3 & 5 / 1.6 & 3 / 0.8 & 2 / 0.8 & 12 / 1.9 \\
    ~\textbf{\ours~(ours)} & 5 / 1.8  & 4 / 2.0 & 4 / 1.9 & 10 / 3.6 & 10 / 3.7 & 8 / 3.1 & 10 / 2.5 \\
    \midrule
    \midrule
    & \multicolumn{7}{c}{\textbf{SFM ground truths \cite{brachmann2021limits}}} \\
    ~MS-Transf. \cite{shavit2021ijcv} & 11 / 6.4 & 23 / 11.5 & 13 / 13.0 & 18 / 8.1 & 17 / 8.4 & 16 / 8.9 & 29 / 10.3 \\
    ~DFNet \cite{chen2022dfnet}  & 3 / 1.1 & 6 / 2.3 & 4 / 2.3 & 6 / 1.5 & 7 / 1.9 & 7 / 1.7 & 12 / 2.6 \\
    ~NeFeS \cite{chen2023refinement}  & 2 / 0.8 & 2 / 0.8 & 2 / 1.4 & 2 / 0.6 & 2 / 0.6 & 2 / 0.6 & 5 / 1.3  \\
    ~\textbf{\ours~(ours)} & 2 / 0.8  & 3 / 1.4 & 3 / 1.3 & 4 / 1.3 & 5 / 1.6 & 6 / 1.6  & 6 / 2.0 \\
    ~\textbf{(ours)} w. DINOv2 \cite{oquab2023dinov2} & 3 / 0.9 & 4 / 1.8 & 3 / 1.5 & 6 / 1.4 & 7 / 2.1 & 8 / 1.8 & 9 / 2.2 \\
    ~\textbf{(ours)} w. RoMa \cite{edstedt2023roma} &  2 / 0.7 & 3 / 1.2 & 2 / 1.0 & 3 / 1.1 & 4 / 1.0 & 5 / 1.4 & 6 / 1.5 \\
    \bottomrule
\end{tabular}
\end{adjustbox}
    \end{center}
    \vspace{-0.1cm}
    \caption{\textbf{Indoor localization}. Indoor scenarios are challenging for our algorithm. Despite this, we achieve competitive results.
    }
    \vspace{-0.1cm}
    \label{tab:indoor-loc}
\end{table}

\myparagraph{Large Scale Localization}
\cref{tab:large-loc} reports results on the large scale benchmark of Aachen v1.1 \cite{zhang2021reference, sattler2012image, sattler2018benchmarking}. The objective of these experiments is to demonstrate the applicability of our method in this scenario in which the previously considered competitors in \cref{tab:small-loc}, namely pose regressors and refiners based on implicit fields, fail to scale. In this setting we mainly compare against PixLoc \cite{sarlin21pixloc}, which is another refinement methods based on a similar idea to our \textit{render\&compare} framework, with a feature-metric error.
We first report results starting from retrieval initialization, showing that our algorithm performs better (except on the finer threshold for Day queries), despite PixLoc trains end-to-end specialized features for localization. In the Supp. Mat. we further discuss trade-offs and similarities of our method with PixLoc, as well as computational cost. 
\\
Additionally, the table shows how our method can complement sota matching-based methods from hloc \cite{sarlin2019coarse}. In this setup, we first run localization using the hloc pipeline. 
We use the estimated poses as an initialization for our method. Results show that we are able to obtain more accurate poses with just 5 refinement steps, adding little overhead. 
\\
\cref{tab:preproc} reports another use-case for our method. Recently, MeshLoc \cite{Panek2022ECCV} has shown a localization pipeline for matching methods using rendered images and a mesh. Depth maps are used to lift the 2D-2D matches to 2D-3D, instead of relying on the SfM point cloud.
We use our method to refine the initial estimate from retrieval; in this way we can provide our refined poses as initialization to a more accurate localizer.  The table shows that, starting from our refined poses, it is possible to achieve a boost in performances while reducing the number of top-K candidates considered.

\myparagraph{Indoor Localization}
On 7scenes \cite{shotton2013scene}, as for Cambridge Landmarks, we compare against method that train on each scene. Indoor scenarios are more challenging for our method, since it is common to have repetitive, textureless surfaces (e.g. walls, floor), which don't provide a meaningful signal for \textit{perceptual similarity}. 
% Additionally, given the limited size of the scene, methods that train on the database images have an additional advantage. 
Despite this, we achieve comparable performances, at the cost of increasing the number of iterations. 
Another factor that affects the evaluation on 7scenes is Ground Truth (GT) accuracy. \cite{brachmann2021limits} demonstrated that the original DSLAM labels are inaccurate, and relased an updated version, named SFM labels. On these more accurate GTs, our results are more competitive. We also test our approach with DINOv2 \cite{oquab2023dinov2}, which leads to comparable precision wrt ImageNet features. This is due to the fact that ViT-based models have a fixed patch size and thus coarser, less-localizable features. To this end, we test the approach from RoMA \cite{edstedt2023roma}, which refines DINO features, and found that they surpass or match other specialized pose refiners. 
However, RoMA was trained for feature matching, an essential step for pose estimation. The improvement suggests that task-specific training can of course improve performance, opening up interesting directions for test-time optimization. More details on these methods are discussed in the Supp. Mat..

\section{Conclusion}
\label{sec:conclusion}
In this work, we investigated whether generic pre-trained features can be transferred to the localization task, thus removing the burden of training dedicated descriptors. Building on the notion that dense feature are robust estimators of \textit{perceptual similarity}, we showed a connection between the latter and \textit{pose similarity}. We demonstrated that this link can be exploited to construct a refinement algorithm within a render \& compare framework, paired with MonteCarlo sampling. 
Experimental results exhibit that our \textbf{\ours}  can be applied in both large and small scenes, either as a standalone refiner or paired with more accurate localizers, and that it can outperform several competitor approaches that optimize dedicated descriptors, especially in outdoor scenarios.

\noindent\textbf{Acknowledgements}
This work was supported by CINI, the European Lighthouse on Secure and Safe AI – ELSA, HORIZON EU Grant ID: 101070617, Czech Science Foundation (GACR) EXPRO (grant no. 23-07973X).
We thank our colleagues who provided helpful feedback and suggestions, in particular Assia Benbihi and Vojtech Panek.

{
    \small
    \bibliographystyle{ieeenat_fullname}
    \bibliography{camera_ready}
}

% WARNING: do not forget to delete the supplementary pages from your submission 
% \input{sec/X_suppl}
\appendix

\section*{Supplementary}
In this supplementary material we show:
\begin{itemize}[noitemsep,topsep=1pt]
    \item an ablation on different scoring functions to demonstrate the effectiveness of simple pixelwise comparison, as mentioned in L500 of the main paper;
    \item a convergence analysis to test the robustness of our algorithm to initialization, as discussed in L422 of the main paper;
    \item additional insights on hyperparameters;
    \item a discussion on inference time;
    \item a pseudo-code version of our algorithm.
\end{itemize}

\section{Scoring functions}
In our paper we showed the effectiveness of dense, pre-trained features for assessing pose similarity, against the previous methods that adopt sparse, specialized features.
The main idea behind these experiments is to test whether dense features provide an actual advantage or if the same results would hold for sparse comparisons as well. Thus, we devised several alternative scoring functions that could be used to rank candidates using sparse comparisons. \\
As a reminder, this scoring function is needed to compute the loss from Eq.~1 of the main paper (
$\mathcal{L_{\mathcal{F}_\theta}}(T | I_q, I_T)$), which at each step is used to compare the rendered candidates against the query, ranking them.
In \cref{tab:abl_score} we compare the following cost functions:
\begin{itemize}
    \item (1): the scoring function adopted in our method; detailed in Eq.~2 of the main paper, \ie, the pixelwise L2 distance between feature maps, normalized along the channels;
    \item (2): a straightforward alternative to dense comparison is to use exhaustive matching of detected keypoints. To this end, we use ALIKED \cite{zhao2023aliked} to obtain for each image a set of keypoints and associated descriptors $\{k_i, f_i\}, k_i \in \mathbb{R}^2, f_i \in \mathbb{R}^d$. Computing the mutual nearest neighbors between the descriptors of the query $I_q$ and a candidate $I_T$, we obtain a set of matched keypoints $\{k_i, k_j\}, i \in K_{I_q}, j \in K_{I_T}$. Finally, the score is the reprojection error among matched keypoints, \ie, their spatial distance (in pixel space). Thus:
    \begin{equation}
        \mathcal{L_{\mathcal{F}_\theta}}(T | I_q, I_T) = \sum_{i \in K_{I_q}, j \in K_{I_T}}{|| k_i - k_j ||^2} \enspace .
    \end{equation}
    \item (3): exhaustive matching increases significantly the cost of computing the loss. Moreover, keypoints that are in opposite locations in the considered image pairs do not provide a useful signal for refining the pose. Thus, a natural alternative to reduce the cost is to match keypoints locally. In this scenario, the nearest neighbors are computed only for keypoints that satisfy $|| k_i - k_j||_2 \leq W$, where $W$ is the patch size that defines the local window around each keypoint in which we compute matches.
    \item (4): implicit matching is an intriguing concept proposed in \cite{cieslewski2018matching}. The main idea is that a standard CNN can be used to extract keypoints, in place of a dedicated keypoint detector. The assumption is that in such networks, each channel has learnt to detect a certain kind of features; thus by looking for local maxima in each channel of the feature maps, these spatial location can be compared among pairs of images without matching descriptors.
    To test this approach, given a feature volume $F_l \in \mathbb{R}^{C, H, W}$, we compute, for each channel: $k_c = \underset{h,w \in H,W}{argmax}\ F_l^{c,h,w}$. They are extracted both for the query $k^q_c$, and a candidate $k^T_c$. \\ 
    To reduce noise we smooth these locations by applying a gaussian filter over a window of size $W$ and then compare them:
    \begin{equation}
        \mathcal{L_{\mathcal{F}_\theta}}(T | I_q, I_T, l) = \sum_{c \in C_l}{|| k^q_c - k^T_c ||^2} \enspace .
    \end{equation}
\end{itemize}

\begin{table}[th]
    \centering
    \begin{center}
    \begin{adjustbox}{width=0.8\columnwidth}
\begin{tabular}{lccc}
    \toprule
    \multirow{2}{*}{Scoring function} & \multirow{2}{*}{ShopFacade} & \\
    && \\
    \midrule
    (1) Dense Comparison  & 12 / 0.45 \\
    (2) Exhaustive Matching &  20 / 0.93 \\
    (3) Patch-wise Matching  & 34 / 1.36 \\
    (4) Implicit Matching  & 85 / 1.92 \\
    \bottomrule
\end{tabular}
\end{adjustbox}
    \end{center}
    \vspace{-0.2cm}
    \caption{\textbf{Ablation on scoring function}. Shows the effectiveness of densely comparing feature maps against more sophisticated cost functions. Median errors reported in $cm /  \degree$.
    }
    \vspace{-0.2cm}
    \label{tab:abl_score}
\end{table}

\begin{table}[th]
    \centering
    \begin{center}
    \begin{adjustbox}{width=\columnwidth}
\begin{tabular}{ll|ccccccc}
    \toprule
    \multirow{2}{*}{Coarse Features} & \multirow{2}{*}{Fine Features} & \multirow{2}{*}{ShopFacade} & \multirow{2}{*}{OldHospital}\\
    &&& \\
    \midrule
    \textit{CNN features: ResNet-18} & & \\
    ~CosPlace \cite{berton2022cosPlace} &  ImageNet &
    12 / 0.45 & 39 / 0.73 \\
    ~ImageNet  & ImageNet & 12 / 0.55 & 46 / 0.80 \\
    ~SimCLR \cite{chen2020simple} & SimCLR \cite{chen2020simple} & 18 / 0.62 & 50 / 0.83 \\
    \midrule
    \textit{Transformer: ViT small} & & \\
    DINOv2 \cite{oquab2023dinov2}   & DINOv2 \cite{oquab2023dinov2}   & 34 / 0.81 & 59 / 1.15 \\
    \bottomrule
\end{tabular}
\end{adjustbox}
    \end{center}
    \vspace{-0.3cm}
    \caption{\textbf{Ablation on feature extractors}, to test whether the property of dense features being robust estimators of \textit{visual similarity}, which is traditionally associated with feature maps from CNN architectures, holds for state-of-the-art vision transformers. Median errors in $cm / \degree$.
    }
    \vspace{-0.2cm}
    \label{tab:abl_dino}
\end{table}

\myparagraph{Results}
Results in \cref{tab:abl_score} show that among scoring functions based on sparse comparisons (2, 3, 4), performances are proportional to the computational cost, \ie, the more accurate is (2) which is also the more expensive.
Overall, simple dense comparison (1) is the best performing one, while being also lightweight and hyperparameter-free. This effect can be understood in light of the discussion in Sec. 3.1 of the main paper: comparing dense features allows to fully exploit the properties of deep networks as \textit{perceptual similarity} \cite{zhang2018pips} estimators, and it provides a smoother signal (w.r.t. sparse features) thanks to the spatial structure of feature maps. \\
This property of dense feature maps was one of the core ideas behind our paper. While this effect was studied mainly with feature maps from CNN architectures \cite{amirshahi2016image,kim2017visual,zhang2018pips}, in \cref{tab:abl_dino} we experiment with the state-of-the-art vision transformer trained in DINOv2 \cite{oquab2023dinov2}. In these architectures each image patch is encoded and processed as a token. In order to use this model for our algorithm, we compute the distance between corresponding tokens in a pair of images, using different layers of the encoder to preserve our \textit{Coarse-to-Fine} approach. \\
While these tokens, paired with positional encoding, preserve spatial information, we find that using these features yields only adequate results, much lower than what can be achieved with a simple ResNet-18. These findings can be explained in light of the receptive field of each token being constrained to be equal or higher than the patch size (14 pixel specifically), and the fact that the self-attention scheme embeds some global context into each patch.
This argument was recently sustained in RoMa \cite{edstedt2023roma}, which proposes to refine DINOv2 features with a specialized CNN architecture.
In Tab.5 of the main paper, we experiment with this architecture and find that it surpasses or match other specialized pose refiners, as shown in the table.  
Note that RoMa features rely on an architecture with roughly $80$x more parameters than the ResNet-18 that we adopt. 

\section{Optimization hyperparameters}
In this section we provide additional insights and ablations on some key hyperparameters of our algorithm. As discussed in Sec. 4.1 of the main paper, a key element for the success of our pose refinement is exploiting a \textit{Coarse-to-Fine} approach, where we gradually move from deeper features to shallower ones. Given that we employ 3 different feature levels (coarse-medium-fine), this entails choosing 2 hyperparameters, namely $N_1$ and $N_2$. $N_1$ indicates after how many steps we switch from coarse to medium; $N_2$, reported as a negative value, represents that the last $N_2$ steps are carried out with the shallower features. \\ 
\cref{tab:abl_N} reports results on 2 Cambridge scenes, showing the effect of these 2 values. For these experiments, when varying $N_1$, we keep fixed the number of steps after $N_1$. When changing $N_2$, the number of steps before is fixed. \\

\begin{figure}[t]
    \begin{center}
    \includegraphics[width=\columnwidth]{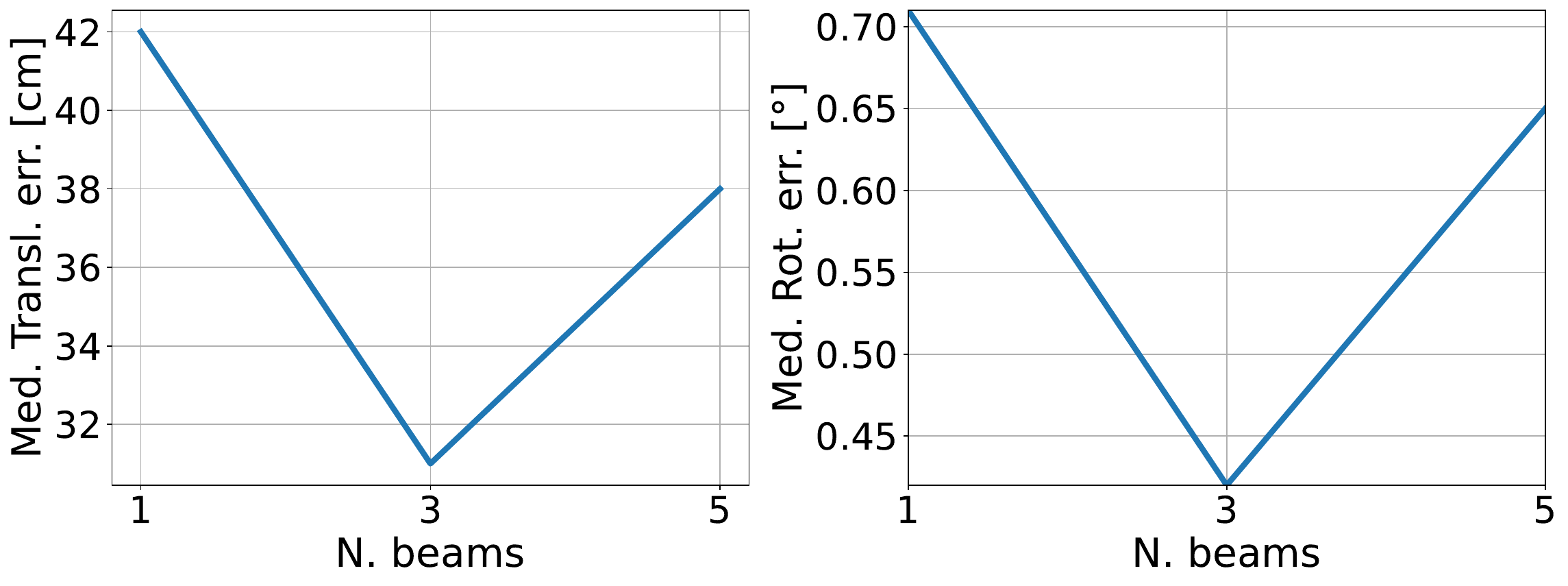}
    \end{center}
    \vspace{-0.2cm}
    \caption{\textbf{Number of independent beams}. Results on KingsCollege.
    }
    \vspace{-0.4cm}
    \label{fig:beams}
\end{figure}

Another important part of our method is multi-hypothesis tracking \cite{choi2012beams}, optimizing independently multiple \textit{beams}. In principle, using more beams should always improve results, although this assumption does not hold if the total number of candidates sampled at each step is fixed, which is desirable in order to contain computational cost. Thus, we study this trade-off in \cref{fig:beams}, where we ablate the effect of not using beams at all (\ie, $n beams = 1$), or more. In our main experiments we use 3 beams. The number of candidates is 50 in the beginning, and it is slowly reduced to 20 in the last steps.

\begin{table}[th]
    \centering
    \begin{center}
    \begin{adjustbox}{width=0.9\columnwidth}
\begin{tabular}{cc|ccccccc}
    \toprule
    \multirow{2}{*}{$N_1$} & \multirow{2}{*}{$N_2$} & \multirow{2}{*}{ShopFacade} & \multirow{2}{*}{OldHospital}\\
    &&& \\
    \midrule
    15 & -10 & 16 / 0.74 & 50 / 1.42 \\
    30 & -10 & \underline{12} / \underline{0.45} & \underline{39} / \underline{0.73} \\
    50 & -10 & 13 / 0.47 & 41 / 0.74 \\
    \midrule
    30 &   0 & 20 / 0.50 & 43 / 0.80 \\
    30 & -20 & 12 / 0.43 & 37 / 0.72 \\
    30 & -30 & 10 / 0.42 & 36 / 0.70 \\
    \midrule
    \bottomrule
\end{tabular}
\end{adjustbox}
    \end{center}
    \vspace{-0.2cm}
    \caption{\textbf{N. of steps before switching to coarser features}. We test different values of $N_1$ (switch from coarse to mid-level features), and $N_2$ (switch to finer features). Underlined values are the default used in the main paper. Median errors reported in $cm / \degree$.
    }
    \vspace{-0.2cm}
    \label{tab:abl_N}
\end{table}

\myparagraph{Results} 
As we state in the main manuscript, our algorithm is robust to the choice of the values of $(N_1,N_2)$, provided that enough steps are performed with coarse features in the beginning.
This is because, despite the fact that all feature levels exhibit a convex basin around each pose, the convergence basin is narrower for shallower features. 
Since initialization from retrieval can yield large baselines, it is important to rely on coarse features for enough steps to refine the pose just enough to fall into the convergence basin of the next finer levels. \\
This is apparent from \cref{tab:abl_N}, as it shows that doing too few steps with \textit{conv3} features ($N_1 = 15$) has the biggest impact on performances. On the other hand, doing more steps does not harm performances, although it makes convergence slower. \\
Regarding the value of $N_2$, doing more steps improves results, however the improvement is small, and for this reason we kept it to $-10$ to exploit the best trade-off between cost and performance gain. \\

\cref{fig:beams} proves the usefulness of relying on multiple optimization threads (\textit{beams}) in parallel.
However, using too many beams is also counterproductive; since the number of candidates is the same in these experiments, if the number of beams increases, each beam will sample less candidates, thus reducing their ability to explore the state space, and ultimately harming performances.

\subsection{Convergence analysis}
As with any refinement algorithm, the accuracy of the initial poses is a crucial factor that affects convergence speed, as well as performances. To study the sensitivity of our method to the initial error, we perform the following experiment, similarly to \cite{zhang2021reference}: we randomly perturb the ground truth poses with different error magnitudes, and then run our algorithm for a fixed number of steps. We use magnitudes of $1, 5, 10, 15$ meters and $5, 10, 20, 30$ degrees, and for each magnitude we repeat the sampling 10 times to carry out a more robust analysis. \\
These experiments are performed on ShopFacade, and we run our optimization for 40 steps. Note that results in the main paper for Cambridge scenes are obtained with 80 steps; in this setup we used less iterations  due to the high number of combinations and repetitions of each experiment (160 runs in total). The emerging trends and the conclusions hold nonetheless.

\begin{figure}[th]
    \begin{center}
    \begin{minipage}{.48\linewidth}
        \includegraphics[width=\textwidth]{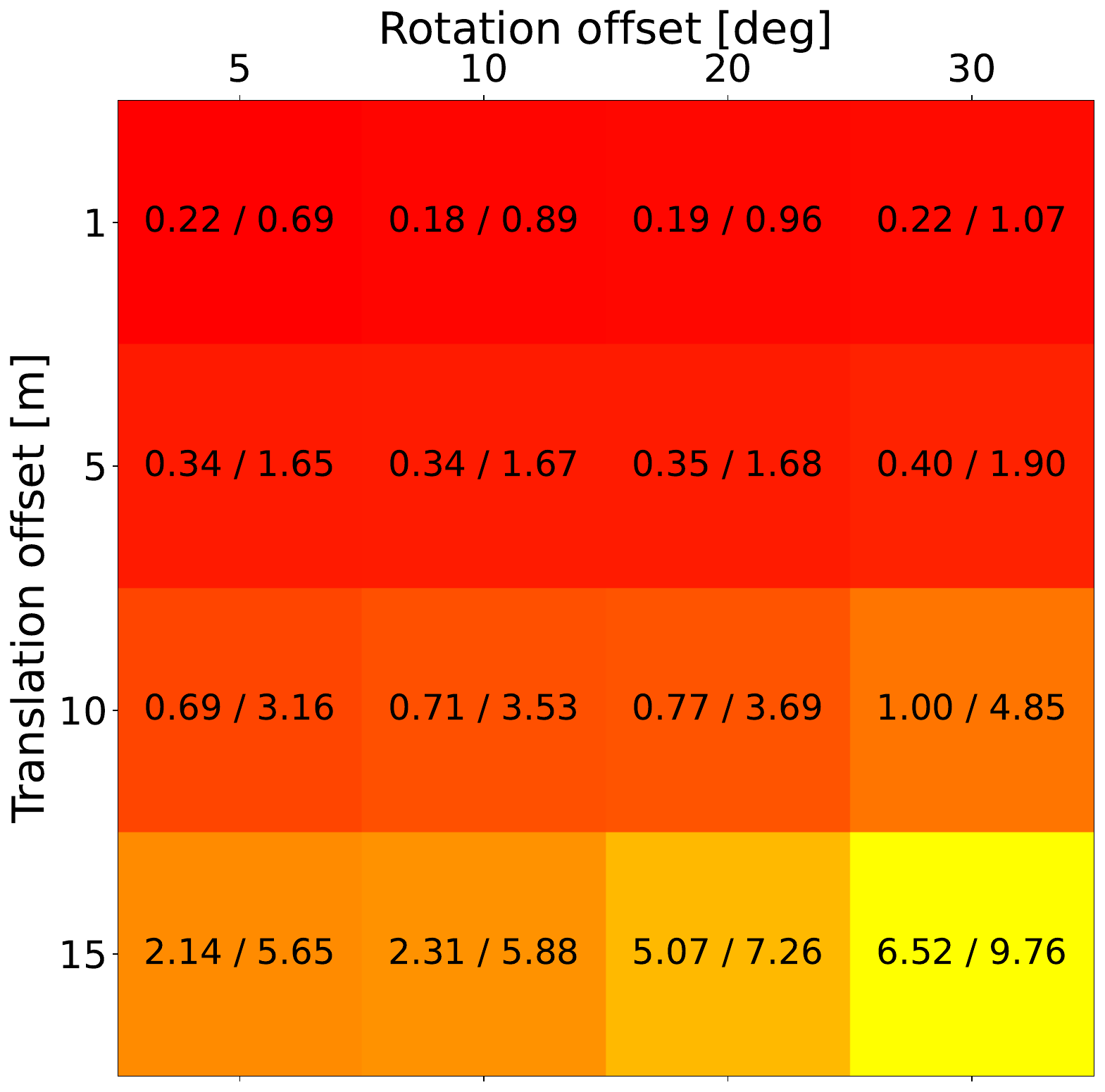}
        \caption{After 20 steps}
    \end{minipage}
    \begin{minipage}{.48\linewidth}
        \includegraphics[width=\textwidth]{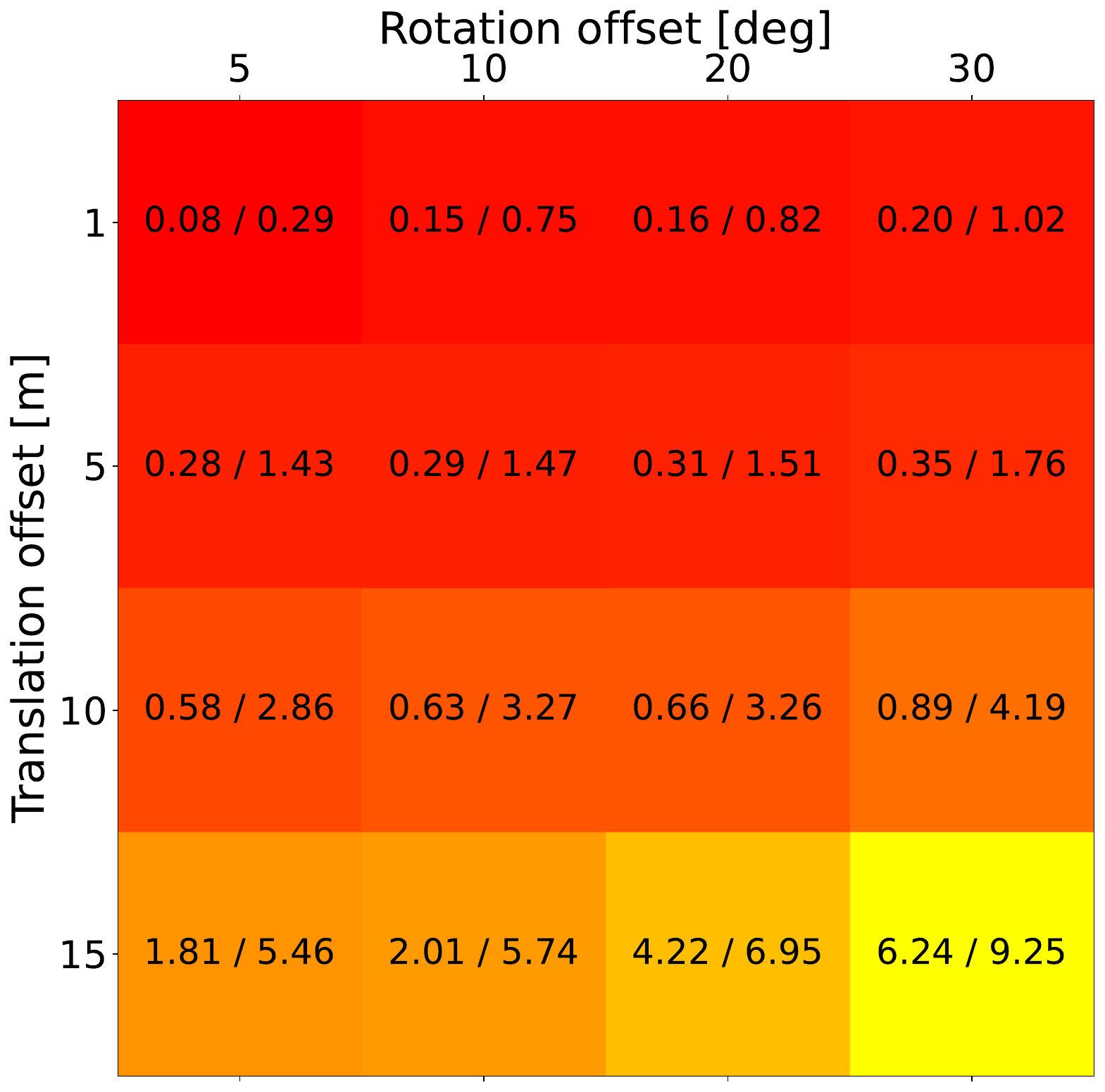}
        \caption{After 40 steps}
    \end{minipage}
    \end{center}
    \vspace{-0.3cm}
    \caption{\textbf{Convergence analysis.} For the scene of ShopFacade, we randomly perturb ground truth poses with fixed magnitudes of error, and then run our optimization to asses its robustness to initialization. Numbers are reported as median errors, averaged over 10 runs, as $m / \degree$.}
    \label{fig:conv_an}
\end{figure}

\myparagraph{Results}
\cref{fig:conv_an} displays in a matrix the results for each translation/rotation combination of errors, after 20 and 40 iterations. At a glance, it is evident from the color map that the obtained accuracy is more correlated with translation error. This effect is understandable as details of the scene might be less recognizable from a distance, thus falling outside of the convergence basin. On the other hand, at a close distance, our optimization can recover from a high rotation error even if there is very little overlap in the views. \\
Overall, our algorithm is robust to errors up to 5 meters, regardless of the rotation, while performances start to degrade at 10 meters.

\section{Inference cost}
We do not claim inference speed among the selling points for our method, since in the literature we did not find a reliable comparison on the same hardware among different methods and implementations. Nonetheless, we report here a breakdown of the time required to optimize a pose over 80 iterations, which is the number of steps that we used to obtain results for Cambridge scenes. Times are measured on a RTX4090.
Rendering the Gaussian cloud from \cite{kerbl3Dgaussians} takes $0.8 ms$; and in total we render 2600 candidates for each query over the steps. \\
Extracting features with a truncated ResNet-18, with FP16 precision and batching takes $0.1 ms$ per image at the lowest resolution ($256\times320$). At the highest resolution that we use ($320\times480$), it takes $0.2 ms$. 
Considering that we use 3 beams, 
our approach takes on average $2.4 s$ for Cambridge, and about $8.7s$ for Aachen (as we perform more iterations). Our optimization relies on independent beams, which can be implemented with multiprocessing, reducing runtime respectively to $1.1 s$ and $4.5 s$ on the same hardware.
When used to refine HLoc poses, we use only 5 iterations, which takes as little as 200ms.

\subsection{Comparison with PixLoc}
On our RTX4090, PixLoc takes $3.1s$ per query independently of the scene.
Our method is more versatile as it does not require any training, and it can be coupled with any dense scene representations, whereas PixLoc requires E2E training and a point cloud. Tab.4 of the main paper shows how our method can be useful as an efficient pre-processing steps in the setting proposed in \cite{Panek2022ECCV}, with different kinds of meshes. Our method also works better than PixLoc as a post-processing step on HLoc poses, and on night queries.  
On indoor datasets and small outdoor scenes, PixLoc achieves superior results, although being slower.

\section{Algorithm pseudocode}

Below in \cref{alg:pseudo} we provide a high-level pseudo-code of our algorithm. 
It highlights: (i) the \textit{render\&compare} structure of our approach, (ii) the fact that the model that we use is a function of the step, (iii) that the particle filter, and thus the noise applied during sampling, are also a function of the step. \\
It does not contain, for simplicity, the multiple beams which are optimized in parallel, or other low-level details.\\
More in detail, the pseudo-code shows, starting from the initial estimate ($est\_center, est\_qvec$), a loop for each query where, in each step: 
\begin{itemize}
    \item The number of candidates (variable $N\_cand$), and the noise magnitude ($noise\_t, noise\_R$) are obtained deterministically as a function of the step;
    \item the particle filter, based on the noise magnitude and current pose estimate, is used to sample $N\_cand$ new hypothesis;
    \item the model is obtained as a function of the step (the backbone will be truncated at a certain layer), and it is used to extract features from the query $q\_feats$ and the sampled candidates $rend\_feats$;
    \item given the features, the sampled candidates are given a score by the function $rank\_poses$; finally the scores are used to update the current estimate
\end{itemize}
We will release our implementation publicly upon acceptance, as we believe it can prove useful to the community.

\begin{algorithm}[th]
\caption{\ours~ pose refinement}
\label{alg:pseudo}
\begin{algorithmic}
\STATE $N \leftarrow n\_steps $
\STATE $renderer \leftarrow load\_scene\_model()$
\FOR{ $query \in query\_list$}
    \STATE $est\_center, est\_qvec \leftarrow init\_pose()$
    \FOR{$step \in 1..N$}
        \STATE $N\_cand \leftarrow get\_N\_cand(step)$ 
        \STATE $noise\_t, noise\_R \leftarrow get\_perturb\_pars(step)$
        \STATE $sampler \leftarrow part\_filter(N\_cand, noise\_t, noise\_R)$
        \newline
        \STATE $poses \leftarrow sampler.sample(est\_center, est\_qvec)$
        \STATE $renders \leftarrow renderer(poses)$
        \newline
        \STATE $model \leftarrow get\_model(step)$
        \STATE $q\_feats \leftarrow extract\_features(query, model)$
        \STATE $rend\_feats \leftarrow extract\_features(renders, model)$
        \newline
        \STATE $center, qvec \leftarrow rank\_poses(q\_feats, rend\_feats)$
        \STATE $est\_center, est\_qvec \leftarrow update(center, qvec)$
    \ENDFOR
\ENDFOR
\end{algorithmic}
\end{algorithm}

\end{document}